# A Machine Vision Method for Correction of Eccentric Error: Based on Adaptive Enhancement Algorithm

Fanyi Wang[1], Pin Cao[2], Yihui Zhang[3], Haotian Hu[1] and Yongying Yang[1*]

*Abstract*—In the procedure of surface defects detection for large-aperture aspherical optical elements, it is of vital significance to adjust the optical axis of the element to be coaxial with the mechanical spin axis accurately. Therefore, a machine vision method for eccentric error correction is proposed in this paper. Focusing on the severe defocus blur of reference crosshair image caused by the imaging characteristic of the aspherical optical element, which may lead to the failure of correction, an Adaptive Enhancement Algorithm (AEA) is proposed to strengthen the crosshair image. AEA is consisted of existed Guided Filter Dark Channel Dehazing Algorithm (GFA) and proposed lightweight Multi-scale Densely Connected Network (MDC-Net). The enhancement effect of GFA is excellent but time-consuming, and the enhancement effect of MDC-Net is slightly inferior but strongly real-time. As AEA will be executed dozens of times during each correction procedure, its real-time performance is very important. Therefore, by setting the empirical threshold of definition evaluation function SMD2, GFA and MDC-Net are respectively applied to highly and slightly blurred crosshair images so as to ensure the enhancement effect while saving as much time as possible. AEA has certain robustness in time-consuming performance, which takes an average time of 0.2721s and 0.0963s to execute GFA and MDC-Net separately on ten 200pixels × 200pixels Region of Interest (ROI) images with different degrees of blur. And the eccentricity error can be reduced to within 10um by our method.

## I. INTRODUCTION

Aspherical optical elements are widely used in large-aperture space telescopes, inertial confinement fusion systems, and high-energy laser systems [1]. Due to some uncontrollable factors in the manufacturing process, some elements' surface will inevitably exist defects, which will not only affect the imaging quality of the system, but also bring great potential risks in industrial application. Therefore, it is necessary to carry out precise detection for surface quality. For the micron-level defects detection requirement of large-aperture aspherical optical elements, a single image can not satisfy both the field of view and the precision of inspection. It is essential to scan the partial images in order. To be specific, to scan the surface of the element through spinning and swinging the mechanical axes, and collect local sub-aperture images, then according to the sub-aperture scanning equation, the acquired sub-aperture images are sequentially stitched [2] to reconstruct the entire picture of the inspected component. The premise of high-precision reconstruction is to adjust as much as possible the coincidence of the optical axis of the element and the mechanical spin axis of the detection system [2]. Low-precision stitching will cause defects to break on the stitched image, misclassification of defect grades and other serious consequences. In view of these problems, this paper proposed an automatic correction method based on machine vision [3], [4] to correct the eccentric error. In this method, when the depth of field of the machine vision system is smaller than the aberration, defocus [5], [6] blur comes into being during imaging. Consequently, defocus restoration for the reference crosshair image is necessary.

Traditional defocus restoration algorithms such as Wiener filter [7]-[9], blind deconvolution [10], [11], least squares filter, and Lucy-Richardson method [12]-[14] all need to know the Point Spread Function (PSF) of the defocus process. Nevertheless, for different aspheric optical elements, their PSFs are different, and are difficult and time-consuming to obtain. Therefore, the traditional defocus restoration algorithms are not applicable here. Given that the defocused crosshair images collected in experiments are similar with the foggy blurred ones, we compared the generation processes of defocus blur and foggy, and found that the mathematical principles of their formations are analogous. Consequently, we innovatively proposed to utilize dehazing algorithm to enhance the grayscale defocused image. Currently, the existing dehazing methods [15]-[19] based on the traditional image process ideas are mainly originated from the dark channel dehazing algorithm [20] proposed in 2009. This algorithm starts from the mathematical principle of the generation of fog, and has excellent performance but high algorithm complexity. The GFA in this paper utilizes guided filter [15] to optimize the soft matting [20] so as to reduce time complexity, but still can not achieve real-time performance. Owing to the rise of convolutional neural networks, a plenty of researchers devote themselves to designing lightweight enhancement networks [21]-[25] to take the place of traditional algorithms so as to achieve a reduction in processing time while retaining effectiveness. For the consideration that for some non-severely defocused images, sacrificing some enhancement effectivity to ensure real-time performance is feasible, therefore the

This research was supported by National Natural Science Foundation of China (NSFC, grant numbers 61627825, 61875173).

The authors labeled 1 are from State Key Laboratory of Modern Optical Instrumentation, Department of Optical Engineering, Zhejiang University, 38 Zheda Road, Hangzhou 310027. Labeled 2 is from Zernike Optics Co., Ltd, 10 Shengang Road, Hangzhou, China. Labeled 3 is from School of Mechatronics Engineering, Henan University of Science and Technology, 263 Kaiyuan avenue, Luoyang, China (e-mail: chuyyy@zju.edu.cn)

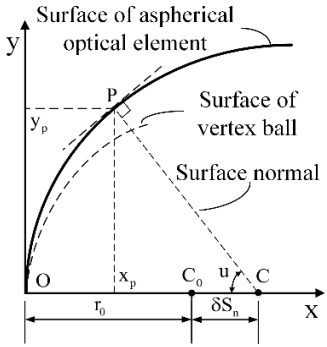

Fig. 1.    Parameters of aspherical optical element.

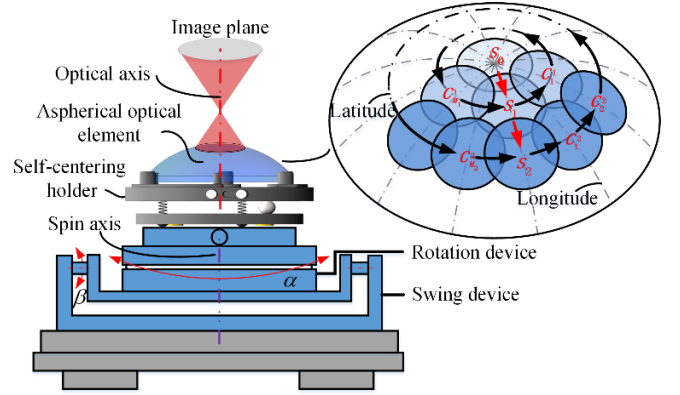

Fig. 2.    Schematic diagram of sub-aperture scanning.

lightweight MDC-Net is designed as a supplement for the poor real-time performance of GFA. AEA is constituted by GFA and MDC-Net, the real-time performance mentioned in the article refers to the average time taken to perform AEA once. Since AEA consists of GFA and MDC-Net, the real-time performance of the two algorithms will be discussed in the experimental part.

Aimed at solving the above two problems of "how to correct the eccentricity error quickly and precisely" and "how to strengthen the defocused crosshair image" during the large-aperture aspheric optical element surface defects detection, combined with deep learning method, an automatic eccentricity error correction method based on AEA is proposed in this paper. First of all, the necessity of eccentricity error correction is presented in Chapter II, and the principle of correction method is explained as well. Then, the eccentricity error correction method is detailly introduced in Chapter III, in which, AEA is mainly introduced, followed by the experiment operation introduction and experiment results analysis in Chapter IV.

## II. Theoretical Background

Aspherical optical elements in this paper refer to those rotationally symmetric curved surface ones. As shown in Fig. 1, the origin $O$ of the coordinate is the vertex of aspherical optical element, there exists a vertex ball whose center $C_0$ is intersected with the optical axis $x$, and the radius of the vertex ball is $r_0$. The aspherical formula is shown as following：

$$z = \frac{c(x^2 + y^2)}{1 + \sqrt{1 - (1+k)c^2(x^2 + y^2)}} + \sum_{i=2}^{N} A_{2i}(x^2 + y^2)^i \qquad (1)$$

Where $c$ is the curvature of the vertex ball which is equal to $1/r_0$, $r_0$ is the radius of the vertex ball, and $k$ is the conic constant. The first term in (1) is a quadratic term, where $k=0$ represents a spherical surface, and this article only considers the case of $k \neq 0$, in other words, aspherical optical elements. The second term is a high-order term.

The surface normal passes through the point $P$, intersects with the optical axis $x$ at $C$. And the normal aberration $\delta S_n$ at point $P$ is the distance between the vertex ball center $C_0$ and the intersected point $C$, which can be calculated by formula (2).

$$\delta S_n = x_p + y_p \cot u - r_0 \qquad (2)$$

In which, $(x_p, y_p)$ is the coordinate of point $P$, $u$ is the angle between the surface normal and the optical axis $x$, and $r_0$ is the radius of the vertex ball.

Fig. 2 illustrates the schematic diagram of the sub-aperture scanning process. As is illustrated, aspherical optical element is clamped by a self-centering holder. For different specifications of elements, the scanning paths are planned according to the surface equations of the elements, which means each moving step $\alpha_i$ and $\beta_i$ of the spin axis $\alpha$ and the swing axis $\beta$ is determined by the surface equation of the element. And then, based on the scanning strategy, sub-aperture image acquisition is conducted along the direction of longitude and latitude of the component surface. The scanning procedure is shown in the upper right corner of Fig. 2. Firstly, capture an image of $S_0$. Secondly, swing a specific angle $\beta_0$ along the longitude direction to reach the position $S_1$, and then refocus and capture the image of $S_1$. Thirdly, rotate the counterclockwise by a specific angle $\alpha_0$ and capture the focused image each time until returns to $S_1$. Then, swing to $S_2$ and repeat the process to perform a missing-free scan of the aspheric surface. After that, the defects detection is performed for each sub-aperture, and finally the distribution of defects on the three-dimensional surface of the component is reversely reconstructed according to the sub-aperture scanning path, and the same defect on adjacent sub-apertures is stitched. The premise of accurate path planning and reverse reconstruction is that the optical axis of the component coincides with the mechanical spin axis. Even the eccentricity error on the order of $100\,um$ between the two axes will lead to the missing collection of partial field of view, the deviation of the reconstruction result at the junction of sub-apertures, and the fracture of defects. Therefore, before the sub-aperture scanning, the eccentricity error between the optical axis of the component and the mechanical spin axis needs to be corrected as much as possible.

According to the imaging characteristic of the aspherical optical element, the crosshair center on the image is equivalent to the center of the optical axis of the element. Under ideal non-eccentricity error condition, the crosshair center coincides with the mechanical spin axis, forming a solid green line image in Fig. 3(b). When there exists eccentricity error, the crosshair center follows the spin motion, and draws a circle around the mechanical spin axis, as shown by the red dotted circle in Fig. 3(b), and the imaging principle is reflected by the red dotted optical path in Fig. 3(a).

The correction principle of eccentricity error used in this article is to fit the trajectory circle of the crosshair centers, and the center of the trajectory circle is the mechanical spin axis theoretically, then control the movement system to move the center $O_c$ of the crosshair image to the location of the mechanical spin axis $O_m$. The key of this method is to accurately extract the pixel coordinate of the crosshair center, but due to the aspheric imaging characteristic, the crosshair image will inevitably show defocus.

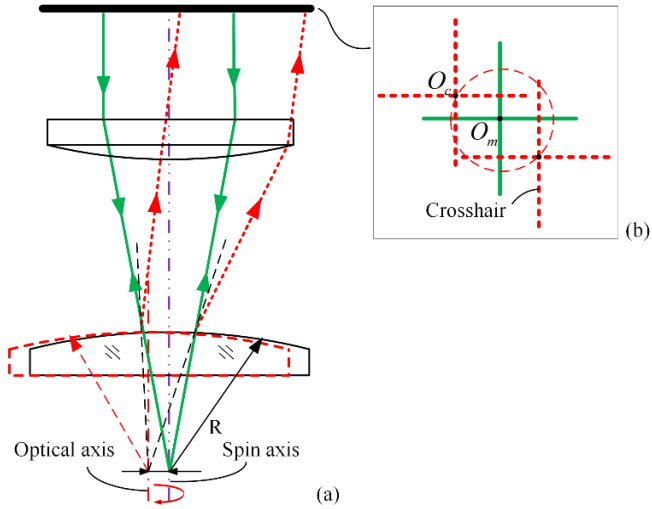

Fig. 3. Optical principle of eccentricity error correction, (a) is the optical path of imaging and (b) is the images of crosshair on the condition of different relative positions of the optical axis and the spin axis.

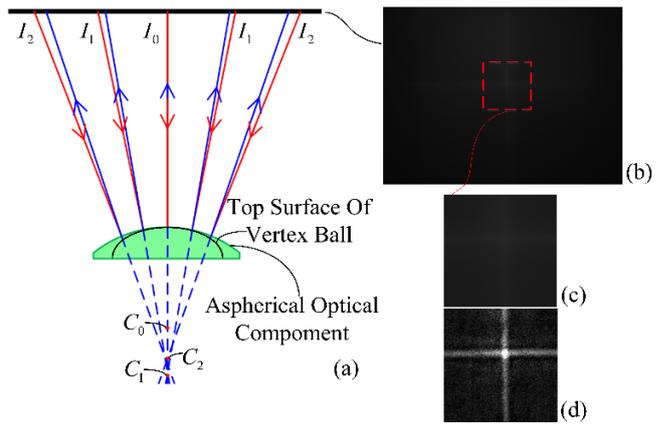

Fig. 4. Aspheric optics vertex ball center imaging simulation, (a) is the ray tracing image, (b) is the defocused crosshair image, (c) is the ROI of crosshair image and (d) is the enhancement result of AEA.

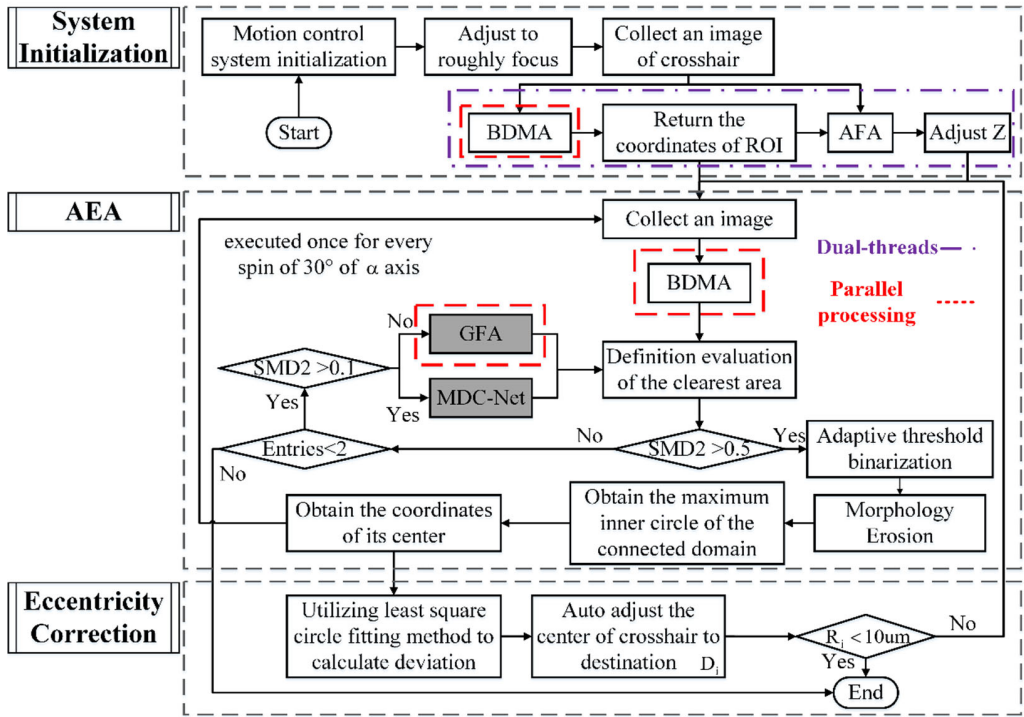

Fig. 5. Flow chart of automatic eccentricity error correction method.

The simulation result in Fig. 4(a) illustrates that due to the aberration, except for the emergent light at the paraxial axis, the others all deviate from the incident light. The incident light $I_1$ and $I_2$ are imaged at point $C_1$ and $C_2$ separately, and the distances between the vertex ball center $C_0$ are their normal aberrations. As can be seen, due to the existence of the normal aberration, the light energy diverges. As is illustrated in Fig. 4(b), the crosshair image formed at the vertex ball center is blurry, with a bright center and dark edge, and the grayscale of background is high which results in that the crosshair center coordinate can not be directly extracted. In view of this problem, our article further proposes AEA to enhance the clarity of the crosshair image. Fig. 4(c) is the ROI of the crosshair image and Fig. 4(d) is the enhancement result of our AEA.

## III. CORRECTION METHOD BASED ON AEA

As is shown in Fig. 5, the automatic eccentricity error correction method can be divided into three steps: System

Initialization, AEA and Eccentricity Correction. The core of our method is AEA, which is composed of GFA and MDC-Net, and they have been deepened out in Fig. 5. The entire eccentricity correction method is optimized and accelerated on both software and hardware, as is marked out in Fig. 5, the purple dotted area uses dual-threads, and the red chain-dotted areas use parallel processing, which greatly improve the efficiency of our method.

The general process of our method is illustrated in Fig. 5. First, implement System Initialization step to make the optical system accurately focus on the vertex ball center of the aspherical optical element, and obtain the pixel coordinate of the ROI where the center of the crosshair is approximately located. Then enter the AEA step, the spin axis spins for one revolution at the setting time, collect a focused image with each rotation angle of $30°$, and evaluate the SMD2 value of the ROI area of each image. If the SMD2 value is greater than the threshold 0.5, the binarization and morphology are directly applied to the ROI to extract the center coordinate of the crosshair image. Otherwise, determine whether the SMD2 value is greater than the threshold value 0.1, and if so, use the faster MDC-Net for clarity enhancement, if not, sacrifice time efficiency and use GFA for clarity enhancement, and then the enhanced image is subjected to the operation of extracting the center coordinate of the crosshair image as described before. Finally, in the Eccentricity Correction step, the least square circle fitting method is performed on the extracted center coordinates of the 12 crosshair images. The center of the fitted circle is the position of the theoretical mechanical spin axis, and then move the crosshair center to this position to correct eccentricity error. To ensure the accuracy of the error correction, we set the constraint condition that the eccentricity error should be less than $10\,um$, loop the AEA and Eccentricity Correction steps until the accuracy requirement is satisfied.

### A. System Initialization

In the System Initialization step, the crosshair is roughly focused on the center of the vertex ball, and collects an image of the crosshair at first. Then starts two threads to respectively execute Automatic ROI Acquisition Algorithm (ARAA) and Auto Focus Algorithm (AFA).

#### A.1 ARAA

It can be draw from the imaging principle that the central area of the crosshair image is the clearest, and can be located by utilizing the definition evaluation function. The definition evaluation function utilized in this article is the normalized SMD2 (it is still called SMD2 for convenience in this paper) in consideration of its excellent sensitivity performance. The expression of SMD2 is as follows:

$$SMD2 = \sum_{x,y=1,1}^{i,j} \frac{|f(x,y)-f(x-1,y)| \cdot |f(x,y)-f(x,y-1)|}{255ij} \quad (3)$$

In which, $i$ and $j$ are the pixel width and height of the image, and $f(x,y)$ is the gray value of the image at pixel coordinate $(x,y)$.

Based on the SMD2 definition evaluation function, we utilize the Block Definition Measurement Algorithm (BDMA) to

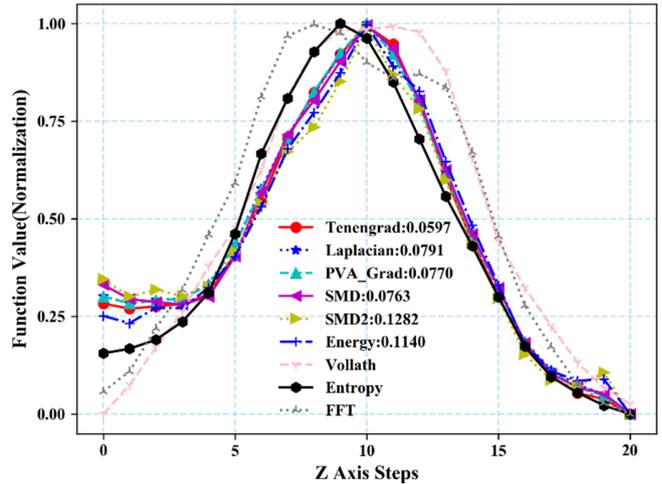

Fig. 6. Curves of 9 types of definition evaluation functions (Tenengrad, Laplacian, PVA Grad, SMD, SMD2, Energy, Vollath, Entropy and FFT). The decimals after the label of the first 6 definition evaluation functions are the differential gradient values at the best focus position.

search for the clearest ROI which contains the crosshair center on the entire image. Next, the system records the top left corner of the ROI so as to locate the relative position of the crosshair center on the entire image. Follow-up processing will only be performed on the ROI which greatly reduces the calculation amount of the whole method. The BDMA is introduced as follows:

First of all, set the width $W$ and height $H$ of the sub-area and the step $T$ which defines the search step to the right or down each time on the input image, and calculate the SMD2 values of all the sub-areas of the specified width and height in parallel on the input image according to the search step. Then, store the SMD2 values in the mapping table data structure in order. Ultimately, arrange the mapping table according to the key values in descending order, and the corresponding sub-region rank at the head of the mapping table is the ROI with the highest definition score on the image, that is, the region where the crosshair center is located, and then returns the coordinate of the upper left corner of this ROI region.

#### A.2 AFA

The acquired image at the beginning is only a roughly focused one, and the comparison of SMD2 values between the images of the adjacent location is required to obtain the best focus position. Set the vertical resolution to $10\,um$, and collect 10 images up and down respectively in step of vertical resolution. Collect 20 images, plus the coarse focus one which is originally acquired, for a total of 21 images. After receiving the end signal from ARAA and the coordinate of the upper left corner of the ROI, calculate the SMD2 values of the ROIs of 21 images, and the one with the largest SMD2 value corresponds to the best focus position.

The key of AFA is to choose an appropriate definition evaluation function which is selected from nine commonly used ones: Tenengrad, Laplacian, PVA Grad, SMD, SMD2, Energy, Vollath, Entropy and FFT [26]. Take twenty-one crosshair images acquired in once experiment as inputs, the above 9 types of definition evaluation functions are calculated and then normalized, their broken lines are shown in Fig. 6, the larger the value, the higher the definition of the ROI corresponding to the abscissa position, and the position with the highest definition can be regarded as the best focus position. In Fig. 6, except for the vertices of Vollath, Entropy and FFT these three broken lines, the vertices of the other six broken lines correspond to the same abscissa and are at the coarse focus position. Therefore, the conclusion can be drawn that the coarse focus position is the best focus position in this experiment. Then, by calculating the differential gradient values of the six broken lines at the best focus position, the definition evaluation function with the highest sensitivity at the best focus position can be selected. The formula to calculate the differential gradient value at $n$ is as follows:

$$\delta =| 2f(n)-f(n-I)-f(n+I)|/2 \quad (4)$$

From the differential gradient values at the best focus position shown in Fig. 6, it can be concluded that SMD2 has the best sensitivity at the best focus position. Therefore, we choose SMD2 definition evaluation function as the evaluation index.

### B. AEA

After the step of System Initialization and the function of the self-centering holder, the eccentricity error between the optical axis of the component and the mechanical spin axis is generally on the order of sub-millimeters. In order to reduce the amount of calculation, only the ROI area is processed. Therefore, the algorithm first uses BDMA to obtain the crosshair center area, and then determines whether its SMD2 clarity index is greater than the prior threshold index 0.5, the threshold is an empirical value obtained by performing a large number of center extraction experiments on images of different clarity. For ROI with SMD2 value greater than the threshold, directly uses the binarization and morphological methods to extract the crosshair center and the steps are as follows:

Firstly, adaptive threshold binarization is performed on the ROI to obtain a binary image, and the mathematical principle of adaptive threshold binarization is as follows:

Determine the size of the binarized single region module as $ksize$ which in our method is 17, and then calculate the gaussian weight value $T(x,y)$ for each pixel in the module.

$$T(x,y)=\alpha \cdot \exp[-(i(x,y)-(ksize-1)/2)^2/(2\cdot\theta)^2] \quad (5)$$

Where $i(x,y)=\sqrt{x^2+y^2}$, and $(x,y)$ is the pixel coordinate with the center of a single $ksize \times ksize$ area module as the origin, $\theta=0.3\cdot[(ksize-1)\cdot0.5-1]+0.8$ and $\alpha$ satisfies $\sum T(x,y)=1$.

The rule of binarization is as follows:

$$dst(x,y)=\begin{cases} 0, src(x,y) > T(x,y) \\ 255, src(x,y) \le T(x,y) \end{cases} \quad (6)$$

Where $dst(x,y)$ is the target binary image, and $src(x,y)$ is the original ROI image.

Secondly, due to there exists noise in the background of the image which results in the stray connected domains on the binarized image, a small structure element is used to perform morphological eroding operation to erase them. The erase result is shown in Fig. 7(a). Based on the characteristics of the bright center and the dim surrounding of crosshair image, we draw the conclusion that the inscribed circle in the center area of the crosshair should be the largest. Hence, the problem becomes searching for the largest inscribed circle in the connected domain and obtain the coordinate of its center, which is equivalent to the crosshair center. The extraction result is shown in Fig. 7(b), and the coordinate position of the crosshair center on the global image can be obtained by combining the coordinate position of the upper left corner of the ROI obtained by BDMA.

For the case where the SMD2 value is less than or equal to the threshold index 0.5, it is necessary to use AEA before extracting the center coordinate of the crosshair image. AEA is composed of GFA and MDC-Net, GFA will be introduced at first.

### B.1 GFA

Previously, the dehazing algorithm was mainly applied to RGB color images. In this paper, the single-channel grayscale defocused image is "dehazed". The flow chart of GFA is shown in Fig. 8.

In computer vision and computer graphics, the generation process of fog can be described by formula (7):

$$I(x,y)=J(x,y)\cdot t(x,y)+A(1-t(x,y)) \quad (7)$$

Where $I(x,y)$ represents a haze image and in this paper stands for a defocused image, $J(x,y)$ is the clear image which requires to be solved, and $A$ is the global atmospheric light condition, which can be analogized to the lighting situation of light source, $t(x,y)$ is the transmittance function, and can be analogized to the optical transfer function here. The acquired image $I(x,y)$ is used as the dark channel image, and the average of the first 1% gray value in the dark channel image is taken as $A$ in this paper, (7) can be simplified to formula (8):

$$\frac{I(x,y)}{A}=\frac{J(x,y)}{A}\cdot t(x,y)+1-t(x,y) \quad (8)$$

Currently, we only know $I(x,y)$, so some prior conditions

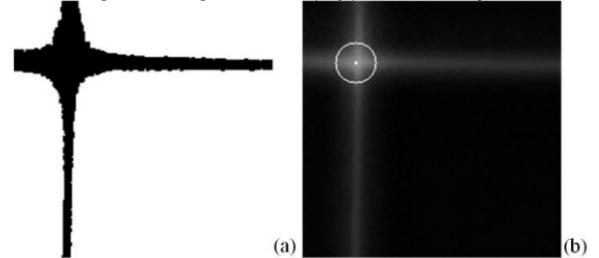

(a)                                    (b)

Fig. 7. Images in the process of crosshair center extraction, (a) is the image after morphological eroding operation and (b) is the extracted center of crosshair.

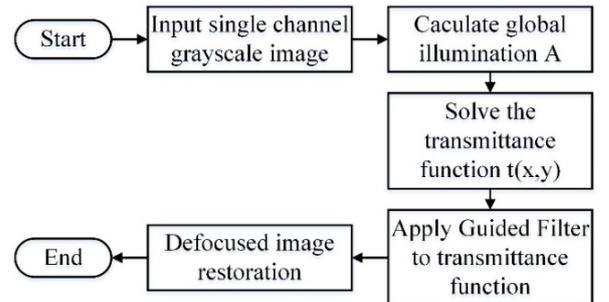

Fig. 8. Flowchart of GFA.

are required for the solution of $J(x, y)$, according to the dark primary color prior theory, we have:

$$J_{\min}(x, y) = 0 \qquad (9)$$

Assume that the transmittance function $t(x, y)$ is locally constant, and take the minimum value on both sides of (9) to get the following formula:

$$\frac{I_{\min}(x, y)}{A} = \frac{J_{\min}(x, y)}{A} \cdot t(x, y) + 1 - t(x, y) \qquad (10)$$

From the prior condition that $J_{\min}(x, y) = 0$, a rough optical transfer function $t(x, y)$ can be obtained:

$$t(x, y) = 1 - \frac{I_{\min}(x, y)}{A} \qquad (11)$$

The boundary of the optical transfer function obtained by the above formula is rough and can not obtain fine enhancement results. Therefore, this paper utilizes guided filter algorithm which has better performance but a little bit slower than fast guided filter with the input image size of $200 \times 200$ to refine the optical transfer function. The time complexity of the guided filter is $O(N)$ comparing to $O(N / s^2)$ of fast guided filter, where $s$ is the scaling ratio of the image, taking 2 in the algorithm, however, the processing time is influenced by many other factors. The mathematical expression of the guided filter is shown in $(12) - (15)$ :

$$\begin{cases} mean_I = f_{mean}(I) \\ mean_t = f_{mean}(t) \\ corr_I = f_{mean}(I.*I) \\ corr_{It} = f_{mean}(I.*t) \end{cases} \qquad (12)$$

Where $mean_I$ and $mean_t$ are the results of the mean filter for $I$ and $t$ respectively, $corr_I$ is the result of self-correlation of $I$, and $corr_{It}$ is the result of cross-correlation between $I$ and $t$.

$$\begin{cases} \mathrm{var}_I = corr_I - mean_I.* mean_I \\ \mathrm{cov}_{It} = corr_{It} - mean_I.* mean_t \end{cases} \qquad (13)$$

$\mathrm{var}_I$ is the variance of $I$, and $cov_{It}$ is the covariance between $I$ and $t$.

$$a = \mathrm{cov}_{It} / (\mathrm{var}_I + \delta), \quad b = mean_t - a.* mean_I \qquad (14)$$

$\delta$ is the regularization parameter.

$$\begin{cases} mean_a = f_{mean}(a, r) \\ mean_b = f_{mean}(b, r) \\ q = mean_a.* I + mean_b \end{cases} \qquad (15)$$

$q$ is the output result after the optical transfer function $t(x, y)$ refined by the guided filter. Substitute the refined optical transfer function $t(x, y)$ into equation (7), the enhanced image $J(x, y)$ can be obtained:

$$J(x, y) = \frac{(I(x, y) - A)}{t(x, y)} + A \qquad (16)$$

### B.2 MDC-Net

Even after optimization, GFA is still problematic in achieving real-time performance in actual implementation, and for some crosshair images with relatively good definition, as shown in Figure. 9(b), certain enhancement effect can be sacrificed to compensate for time efficiency. Based on the mathematical principle of the haze image generation process reflected by (7), we designed a light-weight dehazing network,

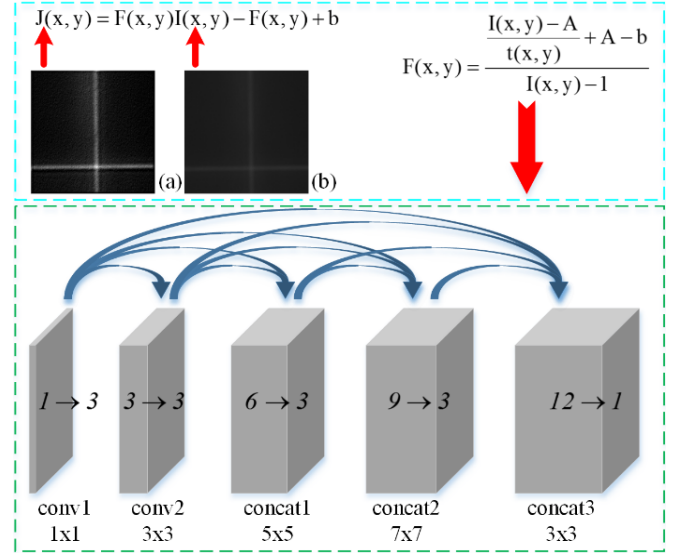

Fig. 9. Structure of MDC-Net, (a) is the enhanced crosshair image and (b) is the origin defocused crosshair image.

called MDC-Net. Formula (17) can be deduced from (7) that:

$$J(x, y) = \frac{I(x, y)}{t(x, y)} - \frac{A}{t(x, y)} + A \qquad (17)$$

In order to integrate two unknown parameters: the lighting situation of light source $A$ and the optical transfer function $t(x, y)$, and one known parameter $I(x, y)$ into one variable, the formula (17) is simplified to (18):

$$J(x, y) = F(x, y)I(x, y) - F(x, y) + b \qquad (18)$$

In which, $F(x, y)$ can be regarded as the equation with the input of $I(x, y)$, and the variables are $A$ and $t(x, y)$, as shown in the following formula (19):

$$F(x, y) = \frac{\dfrac{I(x, y) - A}{t(x, y)} + A - b}{I(x, y) - 1} \qquad (19)$$

Where $b$ is a constant, and the relationship between $F(x, y)$ and $I(x, y)$ can be learned. Therefore, lightweight network MDC-Net is designed to solve this problem. The structure of MDC-Net is shown in Fig. 9.

The whole network is densely connected and contains five convolutional layers. In view of the multi-scale feature extractor is beneficial to the dehazing operation [22], four different scales of convolution kernels are used for feature extraction, and the convolution results are connected in depth. The input of the network is a $200 \times 200$ single-channel grayscale image. After each convolution, the activation function ReLU is applied to increase the nonlinearity of the network.

The network structure is simple but effective, as illustrated in Fig. 9, $conv1$ is a $1 \times 1$ convolution, has 1 input and 3 outputs which is utilized to increase the number of channels and the non-linearity of the network, and to reduce the amount of calculation of the network as well. $conv2$ is a $3 \times 3$ convolution, has 3 inputs and 3 outputs. $concat1$ is the connection of the first two layers in depth, and then the feature map with a depth of 3 is output after $5 \times 5$ convolution. $concat2$ is the connection of the first three layers in depth, and then the feature map with a depth of 3 is output after $7 \times 7$ convolution. The last layer *concat*3 is the connection of all the previous layers in depth, and then the $3 \times 3$ convolution with depth of 1 is used to predict $F(x, y)$. Ultimately, substitute $F(x, y)$ into (18) to obtain the solution of $J(x, y)$.

The network structure is designed in a relatively light form, for the purpose of being real-time. The parameter amount of MDC-Net is only 1.98KB, and the calculation amount is melely 79.12MFLOPs, which is very portable. Although the large-scale convolution occupies a large amount of calculations, it is found through experiments that the multi-scale convolution operation is indeed helpful for the enhancement effect.

### C. Eccentricity Correction

In the previous two steps, the BDMA is executed every time the axis $\alpha$ spins $30°$, and after rotating $360°$, a total of 12 ROIs are collected, and 12 center coordinates of crosshair images are extracted. The theoretical trajectory of the 12 coordinates is a circle, the center of which is the mechanical spin axis. In this paper, the least square circle fitting algorithm is used to locate the fitted circle and obtain the coordinate of the circle center $D_i(X, Y)$ and the radius of the circle $R_i$, in which $D_i(X, Y)$ represents the mechanical spin axis and $R_i$ is equal to the eccentricity error. Next, adjust the center of the crosshair image to $D_i(X, Y)$. Due to the existence of mechanical errors, electrical errors, crosshair center coordinate extraction errors, least square circle fitting errors, etc., it is difficult to satisfy the accuracy requirement by performing eccentricity correction only once. Therefore, set a terminal condition that $R_i < 10um$ for iterative operations for System Initialization step and AEA step is recommended.

## IV. Experiments And Analysis

The experimental test bed is illustrated in Fig. 10, the whole mechanical structure is complicated, and the spin axis $\alpha$ and swing axis $\beta$ involved in this article are marked out.

### A. ARAA Experiment

ARAA is applied to extract the centers of 420 crosshair images. The step T is set to 50 Pixels, and the specified width W and height H are both 200 Pixels. The parameters of 7 samples used in our experiment are shown in TABLE I. Twelve extraction results of ARAA are shown in Fig. 11. Obviously, the centers of the crosshairs are all within the extracted ROIs.

### B. AEA Experiment

The pixel size of the CMOS used in vision system is $5.5um \times 5.5um$, the resolution is $3296 \times 2472$, the magnification $K$ of the machine vision system is 4, and the actual field size is $4.532mm \times 3.399mm$ which can be calculated based on geometric optics knowledge. A parabolic optical element with a vertex ball radius of $18.281mm$ and a conic constant of -1 was selected as the experimental sample. Based on geometrical optics knowledge and known system parameters, depth of field of the system can be calculated, which is 0.0275 $mm$, as illustrated by the solid green line in Fig. 12, and the normal aberration curve is drawn by formula (2), as is shown by the red dashed line in Fig. 12. From the intersection point of the normal aberration and the depth of field, we can draw that for this experimental sample, the imaging area with normal aberrations smaller than the system depth of field is only a small radius area centered on the optical axis, that is, the theoretical none-defocused imaging area is only a circular area with a radius of $1mm$. Meanwhile, due to the surface reflection, internal refraction, and transmission of the optical element, the returned light will inevitably loss a lot, which will further lead to blur.

### B.1 GFA Experiment

After AFA, a crosshair image is acquired. Fig. 13 are the images of the intermediate process of GFA. Fig. 13(a) is the ROI of the original defocused crosshair image. Fig. 13(b) is the distribution image of the optical transfer function obtained by formula (11) which is relatively rough and has obvious graininess. Fig. 13(c) is the optical transfer function smoothed

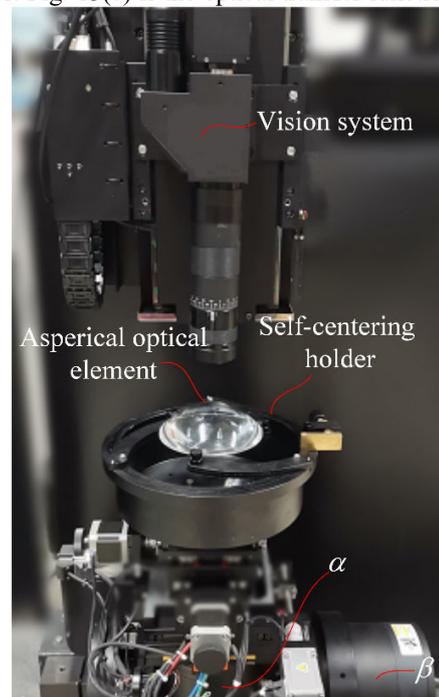

Fig. 10.　Experimental test-bed.

TABLE I
PARAMETERS OF 7 KINDS OF ASPHERIC SURFACE

| $r_0$ / mm | k | $A_4$ / e − 6 | $A_6$ / e − 9 | Number of image |
|---|---|---|---|---|
| 18.2810 | -1.0000 | 2.000 | | 60 |
| 8.8182 | -0.9992 | 86.822 | 63.760 | 60 |
| 13.5510 | -0.6301 | 5.500 | | 60 |
| 13.8590 | -1.0000 | | | 60 |
| 10.9150 | -2.2206 | | | 60 |
| 8.6310 | -7.5380 | | | 60 |
| 31.3840 | -1.9110 | 5.000 | | 60 |

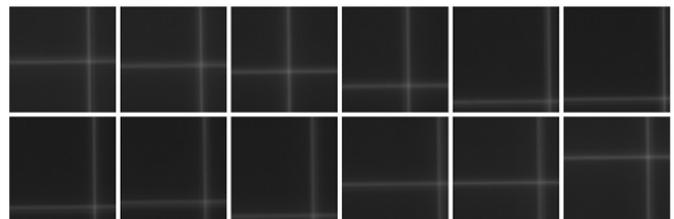

Fig. 11.　The extraction results of ARAA.

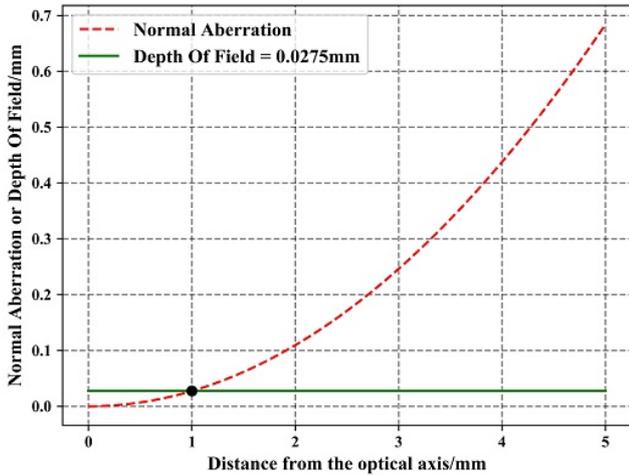

Fig. 12. The relationship between normal aberration and the depth of field of our machine vision system with growing of the distance from the optical axis.

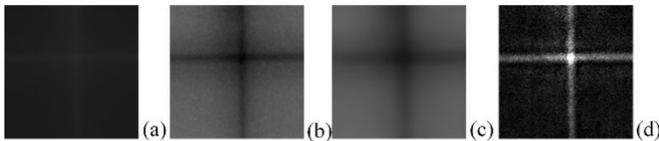

Fig. 13. The image in the process of GFA, (a) is the origin ROI image, (b) is the origin transmittance image, (c) is the transmittance image smoothed by guided filter and (d) is the strengthen result of GFA.

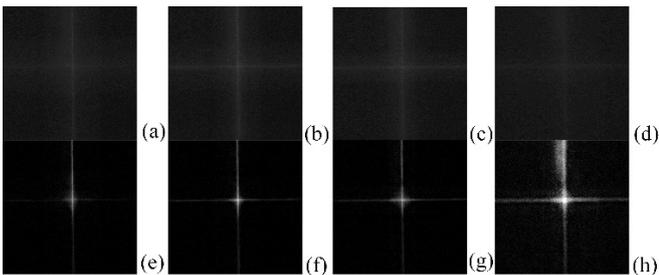

Fig. 14. (a)-(d) are the $500 \times 500$ ROIs of origin defocused crosshair images, and (e)-(h) are the corresponding strengthen results utilizing GFA.

TABLE II
HARDWARE AND SOFTWARE CONFIGURATIONS FOR EXPERIMENTS.

| CPU (for inference） | RAM | GPU(for train) | OS | Environment |
|---|---|---|---|---|
| Intel Xeon CPU E5-2643 v3 @3.4GHz | 64GB | GTX 1080Ti | Windows 10 | Pytorch1.1.0 |

TABLE III
SMD2 VALUES AND PROCESSING TIMES OF 7 KINDS OF ENHANCEMENT ALGORITHMS.

| Algorithm(Network) | SMD2 | Time/s | Avg Time/s | Environment |
|---|---|---|---|---|
| Origin | 0.0024 | | | Python3 |
| Max-Min | 0.4813 | 0.1116 | 0.1204 | Python3 |
| MSR | 0.2557 | 0.0304 | 0.0301 | Python3 |
| **Guided Filter** | **0.9675** | **0.2732** | **0.2721** | C++ |
| Fast Guided Filter | 0.8102 | 0.2313 | 0.2342 | C++ |
| Dehaze-Net[21] | 0.1195 | 0.0781 | 0.8093 | Pytorch |
| AOD-Net[22] | 0.3346 | 0.1094 | 0.1102 | Pytorch |
| **MDC-Net** | **0.7217** | **0.0951** | **0.0963** | Pytorch |

by guided filter. Fig. 13(d) is the enhancement result of GFA. Because the background of the original image has electronic noise during the acquisition process, the enhanced background is grainy, but does not affect the extraction of the center coordinate of ROI.

Select four images with different blur levels, intercept the area where the crosshair center is located and use GFA to perform the enhancement experiment. The original image and enhancement results are shown in Fig. 14. As can be seen, the GFA largely suppresses the background fogging caused by defocusing, the crosshairs are thin and bright, which facilitates the subsequent process of our method. What's more, the success of the GFA experiment confirms our inference and hypothesis.

### B.2 MDC-Net Experiment

In the MDC-Net enhancement experiment, since absolute clear images can not be obtained, and in view of the significant enhancement effect of GFA, the defocused images and the images enhanced by GFA are used to generate input-output pairs for training. The concrete method is as follows, randomly extract $200 \times 200$ ROIs from the center area of the crosshair images and guarantee that the extracted ROI contains the crosshair center by setting constraints and revision, then, apply GFA to those extracted ROIs to make 800 input-output pairs for training, examples of input-output pairs are shown in Fig. 15.

The hardware and software configurations are listed in TABLE II. And for training, the gradient optimization method uses SGD with momentum which is set to 0.99, and train 200 epoches, the initial learning rate is set to 0.0004, and is adjusted by cosine annealing method, for detail, the learning rate is reduced from 0.0004 to 0.00001 in the form of cos function curve. The batch size is set to 20, and a minimum Mean Squared Error (MSE) loss function with "mean" descending mode is used. The constant term $b$ is set to 1.0. The training process uses a GTX 1080Ti GPU. In order to save the time of uploading and downloading images between memory and video memory, inference and subsequent experiments are all performed on CPU. The inputs and prediction results of the trained MDC-Net are shown in Fig. 16. The processing time of GFA is longer than that of MDC-Net. GFA and MDC-Net are both running on the CPU which means there is no upload and download of images between memory and video memory, so the time performance of AEA is better than GFA undoubtedly. If only MDC-Net is used for enhancement, the ROI with severe blur will lead to the overall failure of the method. Under comprehensive consideration, AEA is better than GFA or MDC-Net alone.

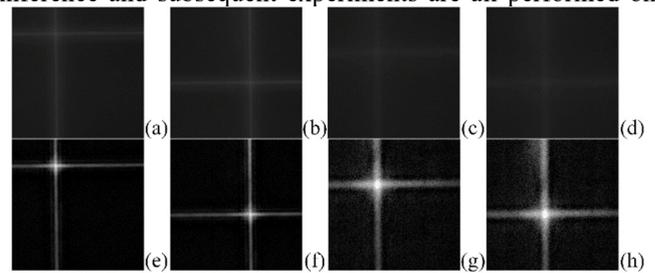

Fig. 15. Training data input-output pairs of MDC-Net.

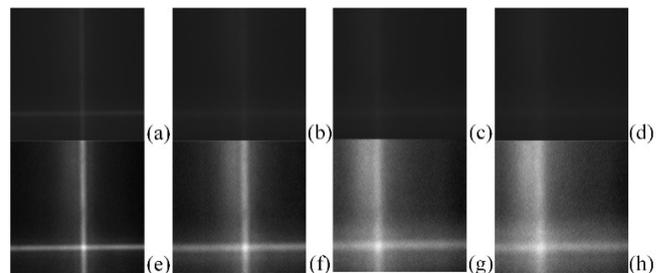

Fig. 16. (a)-(d) are the $200 \times 200$ ROIs of origin defocused crosshair images, and (e)-(h) are the corresponding strengthen results utilizing MDC-Net.

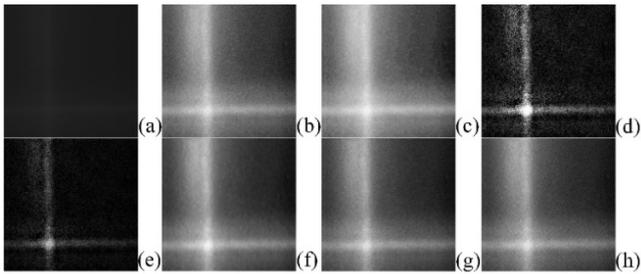

Fig. 17. Comparison of results of four image enhancement algorithms and three networks, (a) is the $200 \times 200$ ROI of origin defocused crosshair image, (b) is the result of Max-Min, (c) is the result of MSR, (d) is the result of GFA, (e) is the result of Fast Guided Filter, (f) is the result of Dehaze-Net, (g) is the result of AOD-Net and (h) is the result of MDC-Net.

TABLE IV
TIME-CONSUMING PERFORMANCE OF GFA AND MDC-Net FOR PROCESSING THREE KINDS OF SCALES OF ROI.

| Algorithm(Network) | Avg Time/ms | | |
|---|---|---|---|
| | $200 \times 200$ | $300 \times 300$ | $500 \times 500$ |
| GFA | 272.1 | 452.8 | 1255.2 |
| MDC-Net | 95.1 | 242.3 | 562.1 |

TABLE V
ECCENTRICITY ERROR CORRECTION DATA.

| Iterations | $R_i$ /pixel | Offset of x /um | Offset of y /um | Eccentric error /um |
|---|---|---|---|---|
| First | 429.776 | -55.89 | -279.16 | 284.996 |
| Second | 49.549 | -26.01 | -0.45 | 26.014 |
| Third | 17.859 | 0.80 | 1.48 | 1.682 |

### B.3 Enhancement and Real-time performance Experiment

As can be seen, the defocus blur of Fig. 16(d) is very serious. However, after the enhancement of MDC-Net, the gray gradient of the center and edge of the crosshair image is separated. For the purpose of proving the enhancement effect and real-time performance of GFA and MDC-Net, we use Max-Min, Multi-Scale Retinxt (MSR) [27], [28], GFA, Fast Guided Filter, Dehaze-Net [21], AOD-Net [22] and MDC-Net to enhance Fig. 16(d), take SMD2 as the definition evaluation index, and compare the time efficiency of those seven methods, the results are shown in TABLE III, and the results of GFA and MDC-Net are bolded.

The enhancement results of Fig. 16(d) using four enhancement algorithms and three networks are shown in Fig. 17. Max-Min and MSR increase the grayscale of both the crosshair and background to a degree which result in that they can not be significantly distinguished. GFA and Fast Guided Filter effectively widen the grayscale level of the crosshair and the background, the processing time of them are similar, but the SMD2 value of GFA is better than Fast Guided Filter, which is obviously more favorable. Because there is no clear image without defocus when training networks, the enhanced result of GFA is used as the focused image. Therefore, the enhancement effect of networks is not as good as that of GFA theoretically. Dehaze-Net, AOD-Net and MDC-Net all pull the grayscale of the crosshair and background to a certain extent. Among them, the SMD2 value 0.7217 of MDC-Net is the highest, and has achieved a great improvement compared with 0.0024 of the original image, which is superior to Max-Min and MSR as well. What's more, it reduces the enhancement time from 0.2732s to 0.0951s compared with GFA.

In order to verify that GFA and MDC-Net both have a certain robustness in time-consuming performance, the above 7 enhancement methods were used for experiments on ten $200 \times 200$ ROI images with different degrees of blur, and the average time was calculated and recorded in the penultimate column of TABLE III. The experimental result shows that the average time is close to the time to process a single image (the third-to-last column), that is, the GFA and MDC-Net are robust in time-consuming performance for ROI images with different degrees of blur.

In order to explore the time-consuming performance of the GFA and MDC-Net on the crosshair images with different scales, ten defocused crosshair images were selected, and $200 \times 200$, $300 \times 300$, $500 \times 500$ ROIs containing the crosshair center were intercepted respectively. Then, the average processing time of the two algorithms were calculated, and the results are shown in TABLE IV.

### C. Eccentricity Correction Experiment

The following is an eccentric error correction experiment. Based on the extracted 12 crosshair center coordinates, the position of the mechanical spin center is obtained by least square circle fitting method. The algorithm sets an iteration terminal condition that the eccentricity error must be less than $10 \ um$, TABLE V shows the results of three iterations during an eccentricity correction process. The eccentricity error is corrected from the initial $284.996 \ um$ to the final $1.682 \ um$.

Fig. 18 vividly displays the process of three iterations. Due to the uncontrollable factors such as mechanical errors in the system, the trajectory of crosshair center coordinates extracted during the eccentricity correction process is not strictly circular.

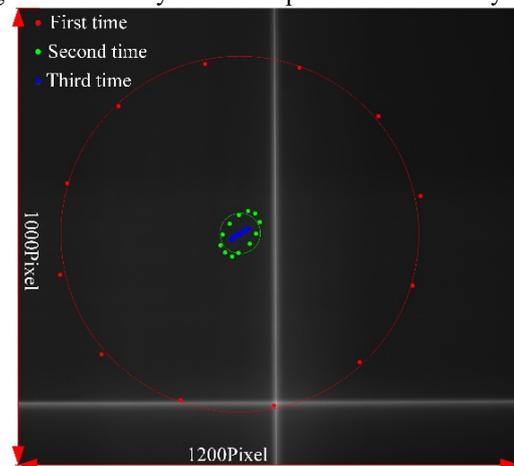

Fig. 18. The process of three times of eccentricity error correction.

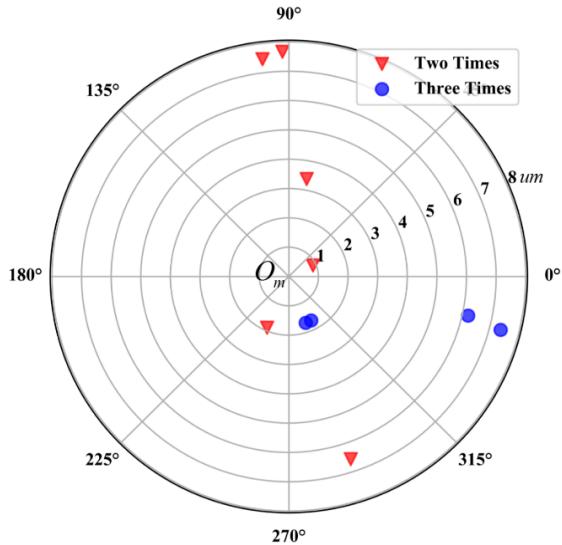

Fig. 19.   Repetitive experiment of eccentricity error correction.

In the first correction, the trajectory of the crosshair centers could be approximately fitted as a circle, which is marked in red. The second time, due to the error, an ellipse is formed, and the trajectory is outlined in green. The third adjustment is close to the mechanical accuracy limitation, and at this time, the trajectory of the crosshair centers are related to the mechanical error. As illustrated in Fig. 18, the radius of three fitted circles are decreasing gradually, and reaching $1.682um$ at last, which is within the terminal condition of less than $10um$.

### D.   Repetitive Experiment

After executing our automatic eccentricity error correction method for 10 times, the final corrected eccentricity errors are plotted in Fig. 19. The entire eccentricity correction algorithm generally requires 2 to 3 iterations. The red inverted triangle in Fig. 19 represents 2 iterations, and the blue circle represents 3 iterations. The origin of the polar coordinate is the center of the mechanical spin $O_m$. The center of the inverted triangle and circle is equivalent to the optical axis position of the aspherical optical element. Therefore, the distance between the scatter point center and the origin represents the magnitude of the eccentricity error. The position of the scatter points in Fig. 19 vividly reflect the relative position of the optical axis and the mechanical spin center of the component after correction. The corrected eccentricity error of 10 times of experiments all can be adjusted to within $10um$, which proves that our correction method is repeatable.

## V.   CONCLUSION

From the experimental results, it can be drawn that ARAA has excellent robustness, and 420 times of repetitive ROI extraction experiments were all successful. The SMD2 value of the defocused image enhanced by AEA has improved by three orders of magnitude than the original one. It respectively takes 0.2732s and 0.0951s to perform GFA and MDC-Net on a $200 \times 200$ defocused ROI image, and the SMD2 values are 0.9675 and 0.7217 separately. And the average time to perform GFA and MDC-Net on ten $200 \times 200$ ROI images with different degrees of blur are 0.2721s and 0.0963s respectively, which proves that the GFA and MDC-Net are robust in time-consuming performance for ROI images with different degrees of blur. After three iterations, the eccentricity error between the optical axis of the aspherical optical element and the spin axis of the mechanical device can be automatically corrected to within $10um$.

This paper solves the scientific research problem of "how to correct the eccentricity error quickly and precisely" during the large-aperture aspheric optical element surface defects detection. And aim at the enhancement for defocused image, this article proposes AEA which is constitute of GFA and MDC-Net. GFA is applied to enhance the defocusing grayscale image base on the analogy between the defocus blur generation model and the dark channel dehazing model. To compensate for the lack of time efficiency of GFA, the MDC-Net is designed based on the mathematical principle of dark channel dehazing model, and ingeniously takes advantage of GFA to make training data set. The machine vision method proposed in this article can adaptively, quickly and precisely realizes the eccentricity error correction of aspherical optical elements, which ensures the accuracy of sub-aperture stitching of large-aperture aspherical optical elements, and provides a more reliable foundation for aspherical surface defects detection.

## ACKNOWLEDGMENT

The authors would like to thank the Associate Editor and the Reviewers for their constructive comments.

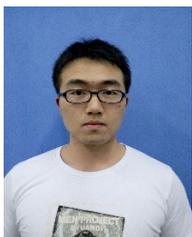

**Fanyi Wang** was born in China in 1995, He received the Bachelor Degree in measurement and control technology and instrumentation from Harbin Institute of Technology, Harbin ,China in 2017, He is currently pursuing Professional Ph.D. Degree in Zhejiang University, Zhejiang, China , majoring in optical engineering. His current research interests are machine vision and precision measurement.

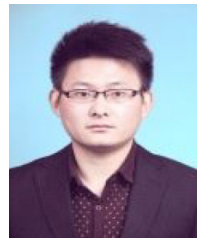

**Pin Cao** was born in China in 1984. He received the Ph.D. Degree in Zhejiang University, majoring in optical engineering. He was invited to be cooperative teacher for graduate students in Polytechnic Institute, Zhejiang University. His current research interests are machine vision, precision measurement and automatic optic inspection of surface defects detection.

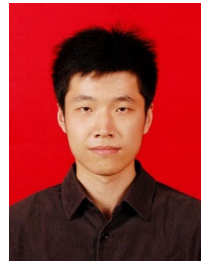

**Yihui Zhang** was born in China in 1992. He received the B.Eng. in measurement & control technology and instrumentation from Changchun University of Science and Technology, Jilin, China in 2014. He received the Ph.D. in optical engineering from Zhejiang University, Zhejiang, China in 2019, and joined the School of Mechatronics Engineering, Henan University of Science and Technology. His current research interests are optical testing and precision measurement.

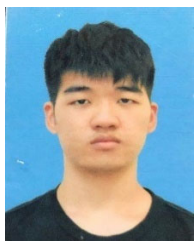

**Haotian Hu** was born in China in 1996, He received the Bachelor Degree in Optoelectronic Information Science and Engineering from Chongqing Normal Universityy, Chongqing, China in 2018 ,He is currently pursuing Master Degree in Zhejiang University, Zhejiang, China , majoring in optics. His current research interest is machine vision.

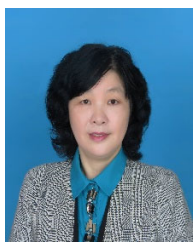

**Yongying Yang** was born in China in 1954, professor, Ph.D , doctoral tutor. she worked at Dept. of Optical Engineering, Zhejiang University. She is now vice chairman of the optical testing professional committee of the Chinese Optical Society. Has won the third prize of national scientific and technological progress, published more than 100 papers. She is mainly engaged in research in this field, digital evaluation and testing technology of super smooth surface defects, interferometry.